\documentclass[preprint]{elsarticle}

\usepackage{times}
\usepackage{amsmath}
\usepackage{booktabs}
\usepackage{natbib}

\journal{Journal of Visual Communication and Image Representation}

\begin{document}

\begin{frontmatter}

\title{A Content-Based Late Fusion Approach Applied to Pedestrian Detection}

\author{Jessica Sena\corref{cor1}}
\author{Artur Jord\~ao\corref{cor2}}
\author{William Robson Schwartz\corref{cor3}}

\address{Smart Surveillance Interest Group\\
	 Department of Computer Science, Universidade Federal de Minas Gerais\\
	   Av. Presidente Ant\^onio Carlos, 6627 - Pampulha, Belo Horizonte, Brazil}

\begin{abstract}
The variety of pedestrians detectors proposed in recent years has encouraged some works to fuse pedestrian detectors to achieve a more accurate detection. The intuition behind is to combine the detectors based on its spatial consensus. We propose a novel method called Content-Based Spatial Consensus (CSBC), which, in addition to relying on spatial consensus, considers the content of the detection windows to learn a weighted-fusion of pedestrian detectors. The result is a reduction in false alarms and an enhancement in the detection. In this work, we also demonstrate that there is small influence of the feature used to learn the contents of the windows of each detector, which enables our method to be efficient even employing simple features. The CSBC overcomes state-of-the-art fusion methods in the ETH dataset and in the Caltech dataset. Particularly, our method is more efficient since fewer detectors are necessary to achieve expressive results.

\end{abstract}


\end{frontmatter}

\section{Introduction}\label{sec::introduction}
Pedestrian detection is an essential task in the computer vision field due to its high demand as input to other tasks such as person recognition,  tracking, robotics, surveillance and transit safety~\cite{Benenson2014Eccvw}. In the past decade, large progress has been made on pedestrian detection, mainly due to two factors: enhancement of features and classifiers. Regarding the former, many studies have shown that the combination of features creates powerful descriptors improving detection~\cite{DollarTPB09,RandForest}. On the latter, ensemble of classifiers have presented an important impact on the accuracy~\cite{Jiang2015CVPR,Correia:2016:ICPR}.

While features improve the human representation, ensemble methods provide considerable enhancements based on the premise that an aggregation of several experts may be superior to a single classifier decision. Regarding this class of methods, there are two main approaches to aggregate the decisions: \emph{early fusion} and \emph{late fusion}. While the first extracts different features and combines them to feed a classifier, the second combines the output of the classifiers by using techniques such as majority voting or weighted voting~\cite{morvant2014majority}.

Late fusion is simpler, presents low-cost and has achieved state-of-the-art results when applied to pedestrian detection~\cite{wagner2016multispectral,Correia:2016:ICPR, pop2017fusion}. In this context, one of the most suitable late fusion methods is the method of Jord\~ao et al.~\cite{Correia:2016:ICPR}, where the authors proposed to fuse detectors based on the spatial consistency between them. They perform the fusion based on the spatial support, i.e., those that agree on having a pedestrian in a certain region. The hypothesis is that a location pointed by multiple detectors has a high probability of actually belonging a pedestrian, while false positive regions have little consensus among detectors (small support) which allows to discard the false positives in these regions.

The issue regarding the false positives in Jord\~ao et al.~\cite{Correia:2016:ICPR} is substantial since the removal of false positive windows is very important but not an easy task. Some false positives are challenging for pedestrians detectors, e.g., a region containing objects with the same vertical orientation of the humans (as trees and lamp posts). False positive windows are intrinsically linked to the feature and classifier used by the detector. Therefore, the diversity generated by a set of detectors might be used, through a late fusion, to tackle the false positive issue, in which the divergence regarding the location of false positives makes possible to perform their reduction.

The approach proposed by Jord\~ao et al.~\cite{Correia:2016:ICPR} aimed at improving the results of a root detector (a input detector which will receive the collaboration of other detectors to increase the confidence in true positive windows), by weighting its windows using spatial support  (spacial consensus) from a set of detectors, thereby, the method is able to reduce the score of false positive windows and increase the score of true positive ones.

Therefore, each time the root detector has a window $w$ overlapping with windows from other detectors, the spacial consensus is used to adjust the score of $w$. The contribution of each detector to $w$ is given by a factor based on the intersection over the union, known as Jaccard (see Equation~\ref{eq::Jaccard} in Section~\ref{sec::methodology}), between its windows and $w$.

Even though Jord\~ao et al.~\cite{Correia:2016:ICPR} achieved accurate results with their simple but effective approach, their technique is not effective in cases where the spatial support is given for a false positive, in these case its confidence will increase and the false positive will be kept instead of being discarded, which would decrease the detection accuracy. To solve this issue, we propose a novel approach to fuse detectors not only based on the spatial consensus among windows but also on the their content.

Our method, referred to as \emph{Content-Based Spatial Consensus} (CSBC), computes the importance (weight) of each detector based on how well-suited it is to make decisions regarding the content inside a particular region of the image before fusing the detectors to adjust the score of a given window. This weight is learned by the Partial Least Squares (PLS)~\cite{Rosipal06overviewand} regression analysis, a dimensionality reduction and regression technique that presents good results in pedestrian detection~\cite{schwartz09c,Melo:2013:ICIP,Melo:2014:ICPR}. 

In our method, the PLS is used to weight windows provided by the detector based on extracted features from the window content according to its label (pedestrian or background). In other words, it estimates which are the characteristics (appearance) that each detector performs better, thereby, we are then able to determinate which are the detectors suitable to each type of content. {For instance, in regions containing trees, it should increase the confidence of detectors that point trees as false positives otherwise, it should reduce the score to this detection during the late fusion. }

To validate our method, we use the widely employed benchmark~\cite{DollarWSP09} in pedestrian detection. We outperform our baseline~\cite{Correia:2016:ICPR} in the ETH dataset~\cite{EssLG07} and achieve comparable results in the Caltech dataset~\cite{DollarWSP09}. Besides, we achieve superior results on Caltech dataset when there are few detectors available to employ the fusion. Additionally, we analyze the influence of the features used to learn the PLS model and conclude that the choice of feature does not affect our approach, which makes the method robust to such factor and enables our method to be efficient even employing simple features.

\section{Related Works}\label{sec::relatedWorks}
Although currently convolutional neural network-based features~\cite{sermanet2013pedestrian,Yang:2016} show state-of-the-art results, proposing features is still a topic of interest and shows good results in pedestrian detection. 
Several feature types have been proposed, including information such as covariance~\cite{paisitkriangkrai2013efficient,tuzel2008pedestrian, paisitkriangkrai2014strengthening}, color~\cite{DollarTPB09, walk2010new, khan2012color, khan2013discriminative}, edge~\cite{dalal:2005,DollarTPB09,lim2013learned,luo2014switchable}, texture~\cite{wang2009hog} and channel-based features~\cite{trichetlbp}. 
The growing number of features at our disposal provides a considerable number of different detectors and detection strategies, suggesting the presence of complementary information among them.

Another topic that has presented interesting results is the fusion of multiple detectors to improve pedestrian detection presenting notable results, mainly when late fusion is applied. 
One of the most successful studies in this context are the works of Jord\~ao et al.~\cite{Correia:2016:ICPR} and Jiang et al.~\cite{Jiang2015CVPR}. The work~\cite{Correia:2016:ICPR} indicates that the employment a set of detectors present a spatial consensus in regions containing a pedestrian, while it diverges in regions containing false positives. Similarly, there are works that propose that pedestrian regions have a strong concentration of high responses, differently from false positive regions, where the responses present a large variance (low and high responses)~\cite{DBLP:conf/ciarp/SchwartzDP11,DBLP:conf/nips/LiSXL10}. Different from Jord\~ao et al.~\cite{Correia:2016:ICPR}, Jiang et al.~\cite{Jiang2015CVPR} combined the responses of multiple detectors to construct a non-maximal weighted suppression called \emph{weighted-NMS}, where it does not discard the windows with lower score (given the Jaccard coefficient), but it uses the score to weight the kept windows.
In their work, the candidate windows (without overlap) are preserved, which might increase the miss rate. This occurs because it is expected that the false positives of distinct detectors reside in different regions of the scene. Therewith, it will not be overlapped and consequently will not be suppressed by the weighted-NMS, keeping the false positive of both the detectors, increasing the miss rate. On the other hand, based on the assumption that the likelihood of different pedestrian detectors generate false positives in the same region is low, Jord\~ao et al.~\cite{Correia:2016:ICPR} showed that removing windows without overlap (i.e., lack of spatial consensus when multiple detectors are considered) improves the pedestrian detection. Besides, Jord\~ao et al.~\cite{Correia:2016:ICPR} showed that the use of multiple detectors provide a strong cue to improve the detection since the spatial consensus for true positives increases the confidence of being a pedestrian while the lack of spatial consensus helps to discard false positive windows.

As showed by Jiang et al.~\cite{Jiang2015CVPR} and Jord\~ao et al.~\cite{Correia:2016:ICPR}, the employment of multiple detectors to improve a single detection has presented notable results. There are also works that employ multiple windows from a single detector to improve its detection precision. The work of Schwartz et al.~\cite{Schwartz:2011:CIARP} is one of the pioneers in this domain. In their work, the authors proposed a novel feature descriptor, \emph{Local Response Context}, which is generated from the responses of an individual holistic detector. Similarly, Ouyang and Wang~\cite{OuyangW13} learn a single detector specialized in multi-pedestrian with a mixture of deformable part-based models. Their work explores the unique visual patterns of multiple nearby pedestrians caused by inter-occlusion and spatial constraint in order to address the problem of detecting pedestrians who appear in groups and have interaction.

In contrast to these approaches, our proposed method does not only use the spatial consensus but also considers the window content to compute the importance of each detector in the ensemble. Thus, in addition to benefiting from the spatial consensus of the fusion, we also weight each detector by considering how suitable it is for a particular region. With this, we are able to eliminate two types of false positives: those where there is a lack of consensus among detectors and those where, despite the consensus, the detectors are known for point out false positives in regions with that particular content.

\section{Proposed Approach}\label{sec::methodology}
We start this section briefly presenting some definitions regarding the work we extend, the Spatial Consensus~\cite{Correia:2016:ICPR}, a late fusion method to perform pedestrian detection. Then, we describe our proposed approach that, in addition to using the premises proposed in the Spatial Consensus, incorporates information regarding the content to pedestrian detection fusion, enhancing the merge performance since it is able to learn for which windows/regions each detector is most suitable.

\subsection{Spatial Consensus}\label{spatial}

As we argued before, the work of Jord\~ao et al.~\cite{Correia:2016:ICPR} is a successful work in the context of late fusion applied to pedestrian detection. Jord\~ao et al.~\cite{Correia:2016:ICPR} combines multiple pedestrian detectors to achieve a more accurate detection.  
Their method, referred to as \emph{Spatial Consensus}, consists of interactively combining windows from different detectors to increase the confidence of candidate windows that actually belong to a pedestrian and discard false positive windows. More specifically, given a set of detectors, the method chooses a detector as root which will have its windows re-weighted based on spatial support (Jaccard coefficient) with the windows of the remaining detectors.
The windows that have no spatial support from other detectors are discarded. The process is summarized as follows: 
\begin{enumerate}
	\setlength\itemsep{0em}
	\item Define the set of detectors $\{det_j\}_{j=1}^n$
	\item Define one detector, called $det_{root}$, to be the input detector, whose the score of its windows will be weighted.
	\item \label{3}  For each window $w_r \in det_{root}$, search for windows $w_j \in det_j$ that satisfy a specific overlap (i.e. spatial support) according to the \emph{Jaccard coefficient} given by
	\begin{equation}\label{eq::Jaccard}
	J_{rj} = \frac{\text{area}(w_{r} \cap w_j)}{\text{area}(w_{r} \cup w_j)}.
	\end{equation}
	\item \label{4}Weight $w_r$ in terms of 
	\begin{equation}\label{eq::weightall}
	score(w_r) =  score(w_r) + \sum_{j=1}^{N} score(w_j)\times J_{rj},
	\end{equation}
	where $N$ is the number of windows given by step 3.
\end{enumerate}
It is important to note that the process described above considers only the spatial support (Equation~\ref{eq::Jaccard}), which we believe is insufficient to reject regions with false positives. Therefore, incorporate information regarding the window content provides a richer clue to discard windows.
Our work proposes to use clues from the window content to weight the scores. As shown in Equation~\ref{eq::weightall}, the Jaccard coefficient is used to re-weight the score of the window $w_r$. This approach faces some problems when a false positive is spatially supported by other detectors, where instead of removing this false positive, the Spatial Consensus increases the confidence of being a pedestrian. To tackle this problem, we propose a weighting approach using Partial Least Squares (PLS)~\cite{wold1985partial} to learn the factor by which the detector will be weighted. To this end, we learn a PLS model for each detector $det_j$, as shown in Equation~\ref{eq::weightpls}, that aims at dynamically penalizing the detectors when they give high scores to false positive windows and reward when they give a high score to an actual pedestrian (i.e., true positive windows). {For instance, the Roerei detector is good for small pedestrians while Franken performs poorly in this type of content. Regarding the false positives, Roerei detects trees as pedestrians while Franken performs poorly with building windows and doors. On the other hand, the LDCF detects false positives in building's facade, trees and poles. Therefore, instead of using only the spatial support information, it is crucial to learn to weigh these detectors based on the semantic information given by the scenario.
	
	To learn the PLS model, we run each detector on a pedestrian dataset and extract features $\theta(w)$ from the windows generated by the detector. Then, we present these features to a PLS and as label we use the Jaccard (Equation ~\ref{eq::Jaccard}) between the window from which this feature was extracted and the closest ground truth window. Thus, in regions containing pedestrians, the PLS model learns what contents the detector succeeds (when it has a high Jaccard with the ground truth) and the types it misses (when it has a low Jaccard). The same happens for the regions with false positives, in which, due to the lack of Jaccard  with the ground truth in these regions, the PLS is able to learn the  background contents that the detector erroneously generates windows (false positives). Thus, the  PLS model created for a detector learns to individually weight any window $w_j$ generated by this detector based on the previous knowledge of the detector for this type of content.
	
	\begin{equation}\label{eq::weightpls}
	score(w_r) =  score(w_r) + \sum_{j=1}^{N} score(w_j)\times PLS_{j}(\theta(w_j))
	\end{equation}
	
	Figure~\ref{fig::pipeline} illustrates the fusion process when a window $w_r$ has spatial support of some of the detectors from $\{det_j\}_{j=1}^n$. For each detector $det_j$, features are extracted from the content of their window ($\theta(w_j)$) and projected in the respective PLS model to estimate the contribution of its score during the combination process. Thus, a detector will only contribute for a window when it is able to achieve good results on windows with such characteristics (this \emph{ability} is learned using PLS in the training stage). For instance, if a detector does not perform well, i.e., provides many false positives in urban regions when a detection window is in urban region, the importance of this detector will be reduced in the fusion for this type of scenes (i.e. yellow detector on Figure~\ref{fig::pipeline}).
	
	\begin{figure}[tbp]
		\centering
		\includegraphics[width=\textwidth]{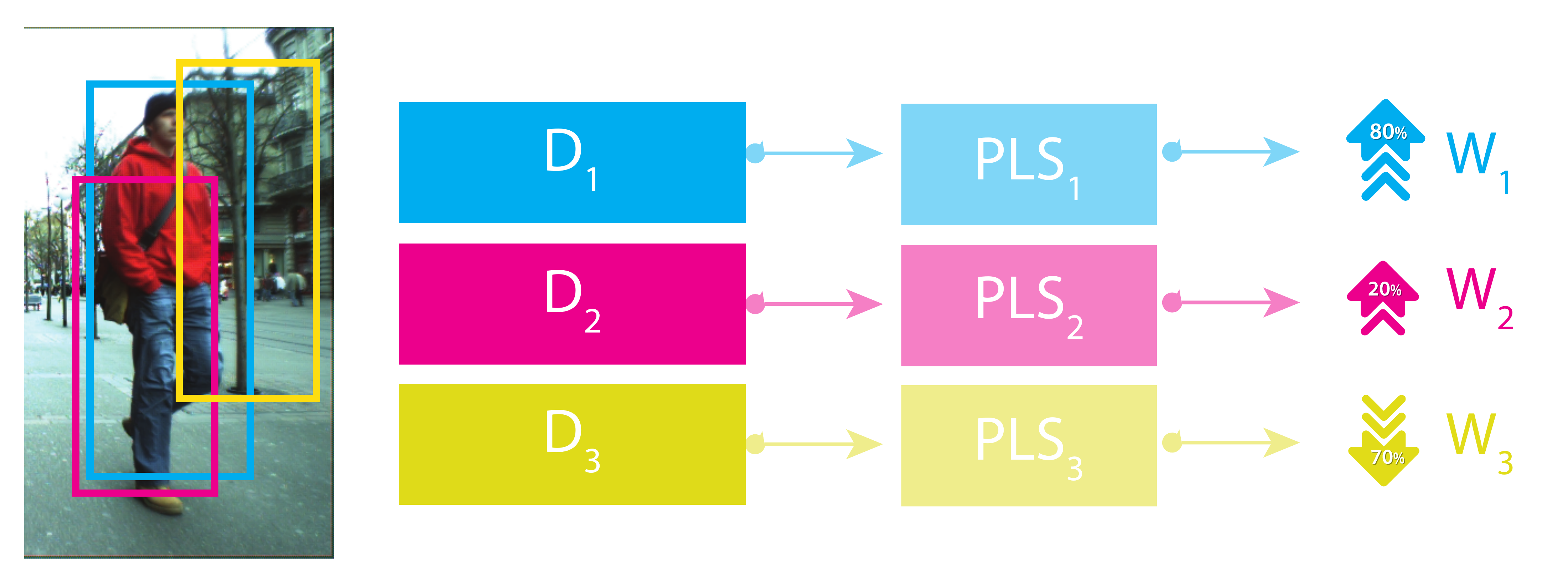}
		\caption{Illustration of the proposed approach. For each detector  (represented by different colors), a PLS model is trained based on the content of the windows provided by the detector during the training. Thus, for a situation where there is space support between detectors, the method considers the contribution of each detector according to its suitability for that type of window. The better-positioned detection window (blue), which has a more pedestrian-like content, is better weighted by PLS. On the other hand, the window whose content has many background characteristics (yellow) has its contribution severely decreased. \textbf{Best visualized in color}.}
		\label{fig::pipeline}
	\end{figure}

\section{Experimental Results}\label{sec::experiments}
In this section we first describe the datasets used to validated our experiments. Then, we present the experimental setup employed, as well as, some details regarding our baseline, the Spatial Consensus~\cite{Correia:2016:ICPR}. Finally, we report and discuss the results obtained.

\subsection{Datasets}
Our experiments were conducted in two pedestrian detection datasets, \textit{ETHZ Dataset} \cite{EssLG07}, and \textit{Caltech Pedestrian Dataset} \cite{DollarWSP09}. Furthermore, the \textit{Inria Person Dataset} \cite{dalal:2005} was used to learn a PLS model for each detector.

\vspace{2mm}
\noindent\textbf{Caltech Pedestrian Dataset.} The Caltech Pedestrian Dataset consists of approximately 10 hours of $640 \times 480$ video taken from a vehicle driving through regular traffic in an urban environment. This dataset provides approximately $50,000$ labeled pedestrians. Moreover, it has been largely utilized by methods designed to handle occlusions since such labels are available.

\vspace{2mm}
\noindent\textbf{ETHZ Dataset.} The ETHZ dataset provides four video sequences, being one for training and three for testing. The images have dimensions of $640 \times 480$, acquired at $15$ frames/second \cite{Schwartz:2011:CIARP}. People height variation and the large pose make this dataset a challenging pedestrian detection dataset.

\vspace{2mm}
\noindent\textbf{Inria Person Dataset.} This dataset provides both positive and negative sets of images for training and testing. The positive training samples are composed from $2416$ cropped images of size $64 \times 128$ pixels, horizontally reflected from $1208$ distinct samples.
There are available 1218 images containing only background that are used to generate negative samples.
The large pose variation, wide range of illumination and weather conditions make this dataset a challenging pedestrian detection dataset. Most methods use \textit{INRIA} to train their detectors. We followed the same procedure, we trained our method using only samples from this dataset.

\subsection{Experimental Setup} 

We are evaluating our results based on the test protocols established in the Caltech pedestrian benchmark \cite{DollarWSP09}, where the results are reported on the log-average miss rate (lower values are better). {We use the code available in the toolbox\footnote{www.vision.caltech.edu/Image\_Datasets/CaltechPedestrians/} of this benchmark to perform our evaluation.}

To perform a fair comparison with our baseline~\cite{Correia:2016:ICPR}, we follow the same protocols used by Jord\~ao et al.~\cite{Correia:2016:ICPR} for setting $det_{root}$ and the set of detectors $\{det_j\}_{j=1}^n$. 
Once specified $\{det_j\}_{j=1}^n$, the order of the set members does not affect the result, since  all detectors of $\{det_j\}_{j=1}^n$ must be evaluated (according to PLS) to discard a window of $det_{root}$ (these question has been noticed by Jord\~ao et al.~\cite{Correia:2016:ICPR}). We also use the same score calibration step describe in Jord\~ao et al.~\cite{Correia:2016:ICPR}. The INRIA Person Dataset is used to acquire the set of scores $\tau$ used to map the response of each detector of $\{det_j\}_{j=1}^n$ to the same response scale of the $det_{root}$. In addition, we report the results of Jiang et al.~\cite{Correia:2016:ICPR}(previously overcome by Jord\~ao et al.~\cite{Correia:2016:ICPR}).

\subsection{Results and Discussion}
In our main experiment, we evaluate the performance of the content-based spatial consensus compared to the our baseline~\cite{Correia:2016:ICPR}. The experiments were performed using Histogram of Oriented Gradients (HOG) features to feed the PLS method. In a complementary experiment, we analyze the robustness of our method regarding the features used in PLS.

\subsubsection{Spatial Consensus vs. Content-Based Spatial Consensus}
Following the experiments of Jod\~ao et al.~\cite{Correia:2016:ICPR}, we evaluate the performance of adding multiple detectors to extract the content-based spatial consensus. According to results shown in Table~\ref{eth}, our method outperforms the baseline throughout the addition of detectors, achieving lower miss rates in the ETHZ dataset. 

The best result of our approach is achieved adding all detectors ($32.12\%$), improving the state-of-the-art (\cite{Correia:2016:ICPR}) in $0.7$ percentage points (p.p.). Table~\ref{caltech} shows that our method overcomes the baseline during the addition of the first four detectors. This is important for two reasons. First, it is that not always possible to have available a large number of detectors to merge. Second, the use of fewer detectors make methods faster. These reasons are common issues when using pedestrian detection methods in the real world. On the other hand, when more detectors were added, we achieved slightly worse results. We believe this occurs because the Caltech dataset is very different from the used training dataset (INRIA). This difference mainly consists of how the image is captured, which leads to a large difference in the content of the windows. {In the INRIA dataset, the images are photos taken in multiples locations while in the Caltech the images are sequences of videos taken from a vehicle driving through regular traffic in a urban environment.}

\begin{table*}[]
	\centering
	\caption[Comparison with baseline (\cite{Correia:2016:ICPR}) in ETH dataset]{ETH Detectors Accumulation. We only accumulate the detectors used by Jord\~ao et al.~\cite{Correia:2016:ICPR} for a fair comparison. The results are measured in log-average miss-rate (lower is better).}
	\label{eth}
	\begin{tabular}{l|c|c|c|l|c|}
		\cline{2-6}
		\multicolumn{1}{c|}{}        & Roerei  & Franken & LDCF    & RandForest & VeryFast \\ \hline
		\multicolumn{1}{|l|}{Jiang et al.~\cite{Jiang2015CVPR}} & $35,19\%$ & $39,69\%$ & $40,93\%$ & $48,76\%$    & $49,32\%$  \\ \hline
		\multicolumn{1}{|l|}{Jord\~ao et al.~\cite{Correia:2016:ICPR}} & $35,63\%$ & $34,61\%$ & $33,89\%$ & $34,15\%$    & $33,98\%$  \\ \hline
		\multicolumn{1}{|l|}{Our}    & $34,22\%$ & $33,19\%$ & $32,47\%$ & $32,46\%$    & $32.12\%$  \\ \hline
	\end{tabular}
\end{table*}

\begin{table*}[]
	\centering
	\small
	\caption{Caltech Detectors Accumulation. We only accumulate the detectors used by Jord\~ao et al.~\cite{Correia:2016:ICPR} for a fair comparison. The results are measured in log-average miss-rate (lower is better).}
	\label{caltech}
	\begin{tabular}{l|c|c|c|c|c|c|c|}
		\cline{2-8}
		\multicolumn{1}{c|}{}        & Roerei  & Franken & LDCF    & Inf.Haar & SCCPriors & R.Forest & W.Channels \\ \hline
		\multicolumn{1}{|l|}{\cite{Jiang2015CVPR}} & 40,59\% & 40,55\% & 48,75\% & 49,65\%       & 45,95\%   & 44,13\%       & 44,68\%      \\ \hline
		\multicolumn{1}{|l|}{\cite{Correia:2016:ICPR}} & 36,90\% & 37,90\% & 27,87\% & 27,11\%       & 24,54\%   & 24,60\%       & 23,67\%      \\ \hline
		\multicolumn{1}{|l|}{Our}    & 27,61\% & 26,53\% & 25,76\% & 25,47\%       & 24,71\%   & 25,07\%       & 24,46\%      \\ \hline
	\end{tabular}
\end{table*}

\subsubsection{Influence of Feature Descriptors}
To evaluate the robustness of our method regarding the feature descriptors used to learn the PLS model, we tested four different features methods: Histogram of Oriented Gradients (HOG), Gray Level Co-occurrence Matrices (GLCM), deep features extracted from VGG16 network~\cite{simonyan2014very} (pre-trained on the ImageNet dataset~\cite{imagenetcvpr09}) and raw gray pixels.

HOG features, proposed by~\cite{dalal:2005}, divides the detection windows into blocks of $16\times16$ pixels with a shift of $8\times8$ pixels between blocks and compute occurrences of gradient orientation. We choose HOG descriptor since this approach shown over the years particularly suitable for human detection in images. 

Gray level co-occurrence matrices (GLCM)~\cite{haralick1973textural} descriptor have been proved to be a very powerful descriptor used in image analysis, since can measure the texture by calculating how often pairs of pixels with specific values and in a specified spatial relationship occur in an image.

Based on the impressive accuracy gains of methods based on Deep Neural Networks in a large number of computer vision tasks (learning features included), we employ a transfer learning technique using a VGG16 network to extract deep features. 

The last feature we analyze was raw gray pixels, the purest way to represent an image since it uses the entire contents of the image justing normalizing the image to grayscale without converting to any representation. We intend to evaluate raw pixels as a proof of concept, to know how the method behaves with a semantically poor feature.

Table~\ref{features} summarizes our method employing the features above-mentioned. There is a slight influence on the choice of the feature. We believe HOG has a better result than GLCM (i.e., approximately 2 p.p.) due to the density of the extraction grid, which promotes a feature vector with more information. To test this hypothesis, we concatenated both descriptors which provided us an even richer descriptor, reducing miss rate in 0.4 p.p. when compared with HOG features. The transfer learning using VGG16 network have a slight worse result when compared to HOG features (we believe this is due to the lack of fine-tuning in our experiment). Finally, we evaluated the robustness of the method regarding a feature considered simple (raw pixels). On this feature, the method proved to be robust, varying slightly (0.07 p.p.) when compared to the most similar feature that calculates co-occurrence of grayscale pixels and varying moderately (2.6 p.p.) when compared to the combination of features (HOG + GLCM) which obtained the best result. Based on the results, we conclude that our method is able to achieve good results even with simple features, therefore being invariant the choice of features.

\begin{table}
	\centering
	\caption{Comparison of the results achieved in the ETHZ dataset by varying the features used to train the PLS. The results are measured in log-average miss-rate (lower is better).}
	\label{features}
	\begin{tabular}[bp]{l|c}	
		\toprule
		Feature Descriptor & log-average miss rate ($\%$)\\ 
		\midrule
		HOG        & 34.22 \\ 
		GLCM       & 36.28 \\ 
		VGG16 & 35.85                      \\ 
		Gray pixel & 36.35                    \\ 
		HOG + GLCM & 33.75 \\
		\bottomrule
	\end{tabular}
	
\end{table}

\section{Conclusions}\label{sec::conclusions}
This work presented a novel late fusion method, called Content-Based Spatial Consensus (CSBC), which is able to weight the contribution of a set of detectors in a fusion based on the information regarding the window content. 
Our approach employs Partial Least Squares  to learn the contribution of a given detector according to how well-suited it is to make decisions regarding the content within the region that the fusion is being made. The proposed method adds intelligence to the fusion of multiple detectors windows which leads to improve the results passing the edge of the previous approach presented in~\cite{Correia:2016:ICPR} on two widely employed datasets, the ETH dataset~\cite{EssLG07} and the Caltech Pedestrian dataset~\cite{dalal:2005}. Additionally, we analyze the influence of the features used to learn the PLS model and conclude that the choice of feature does not affect our approach, which makes the method robust to such factor.

\section*{Acknowledgments}
The authors would like to thank the Brazilian National Research Council -- CNPq (Grant \#311053/2016-5), the Minas Gerais Research Foundation -- FAPEMIG (Grants APQ-00567-14, RED-00042-16 and PPM-00540-17) and the Coordination for the Improvement of Higher Education Personnel -- CAPES (DeepEyes Project).
\section*{References}

\bibliographystyle{model1-num-names}
\bibliography{references}

\end{document}